\pdfoutput=1

\documentclass[11pt]{article}

\usepackage[]{emnlp2021}

\usepackage{times}
\usepackage{latexsym}
\usepackage{graphicx}
\usepackage{subcaption}
\usepackage{xcolor}
\usepackage{colortbl}
\usepackage{latexsym}
\usepackage{amsmath}
\usepackage{multirow}
\usepackage{amsfonts}
\usepackage{algorithm}
\usepackage[]{algpseudocode}

\DeclareMathOperator*{\argmin}{arg\,min}

\usepackage{booktabs}
\usepackage{arydshln}
\definecolor{Gray}{gray}{0.95}
\newcolumntype{a}{>{\columncolor{Gray}}c}

\newcommand{\astfootnote}[1]{
    \let\oldthefootnote=\thefootnote
    \setcounter{footnote}{1}
    \renewcommand{\thefootnote}{\fnsymbol{footnote}}
    \footnotetext{#1}
    \let\thefootnote=\oldthefootnote
}

\usepackage[T1]{fontenc}

\usepackage[utf8]{inputenc}

\usepackage{microtype}

\title{Multilingual Chart-based Constituency Parse Extraction from Pre-trained Language Models}

\author{Taeuk Kim$^*$ \\
  Dept. of Computer Science \\ 
  Dept. of Artificial Intelligence \\
  Hanyang University \\
  \texttt{kimtaeuk@hanyang.ac.kr} \\\And
  Bowen Li \\
  ILCC \\ 
  School of Informatics \\
  University of Edinburgh \\
  \texttt{bowen.li@ed.ac.uk} \\\And
  Sang-goo Lee \\
  Dept. of Computer Science \\
  and Engineering \\ 
  Seoul National University \\
  \texttt{sglee@europa.snu.ac.kr} \\}

\begin{document}
\maketitle

\begin{abstract}
As it has been unveiled that pre-trained language models (PLMs) are to some extent capable of recognizing syntactic concepts in natural language, much effort has been made to develop a method for extracting complete (binary) parses from PLMs without training separate parsers.
We improve upon this paradigm by proposing a novel chart-based method and an effective top-K ensemble technique.
Moreover, we demonstrate that we can broaden the scope of application of the approach into multilingual settings.
Specifically, we show that by applying our method on multilingual PLMs, it becomes possible to induce non-trivial parses for sentences from nine languages in an integrated and language-agnostic manner, attaining performance superior or comparable to that of unsupervised PCFGs.
We also verify that our approach is robust to cross-lingual transfer.
Finally, we provide analyses on the inner workings of our method. For instance, we discover universal attention heads which are consistently sensitive to syntactic information irrespective of the input language.
\end{abstract}

\section{Introduction}

\astfootnote{This work was mainly conducted when TK was at SNU.}
\setcounter{footnote}{0}

Constituency parsing is a classic task in natural language processing (NLP), whose goal is to construct a phrase-structure tree for a given sentence.
As parse trees have long been recognized as being integral to the meaning of sentences, there has been an enormous amount of work in the literature to develop constituency parsers (\citet{charniak-2000-a}; \citet{collins-2003-head}; \citet{petrov-etal-2006-learning}; \textit{inter alia}), resulting in 
the development of sophisticated neural supervised parsers \cite{kitaev-klein-2018-constituency,kitaev-etal-2019-multilingual,zhou-zhao-2019-head}.
Although it becomes possible to obtain such neural parsers of high-quality for a few languages, there remain many other languages which, for lack of resources or attention, have yet to benefit from the progress made in the field of constituency parsing.
The main issue is that it is expensive and time-consuming to prepare adequate numbers of gold-standard trees essential for training parsers with supervision.

Considering this data scarcity problem, \textit{unsupervised} constituency parsing methods have naturally arisen as an alternative for generating constituency trees. 
Work on unsupervised parsing (\citet{shen2018neural,shen2019ordered,kim-etal-2019-compound,kim-etal-2019-unsupervised}; \textit{inter alia}) has focused on devising linguistically-informed neural models, which are carefully designed to be more sensitive to the hierarchical nature of language structure and be able to learn this nature from raw text rather than gold-standard trees. 
Despite the recent progress in implementing \textit{English} unsupervised parsers with decent performance, it has been known as not trivial to reproduce such a success in multilingual environments \cite{kann-etal-2019-neural,zhao2021empirical}.
Moreover, as most unsupervised parsers are typically trained on monolingual text for a specific target language, it is required to prepare separate instances of them to support different languages.

On the other hand, a different line of work \cite{Kim2020Are,wu-etal-2020-perturbed} has proposed a new direction of inducing syntax trees, dubbed in this paper as \textsf{\small \textbf{C}onstituency \textbf{P}arse \textbf{E}xtraction from \textbf{P}re-trained \textbf{L}anguage \textbf{M}odels (\textbf{CPE-PLM})}, by relying on the combination of (i) simple distance metrics and (ii) the representations obtained from pre-trained language models (PLMs).\footnote{We use the term \textit{pre-trained language models (PLMs)} to refer to BERT-like \cite{devlin-etal-2019-bert} Transformer \cite{vaswani2017attention} models that are pre-trained with massive plain text corpora in a self-supervised fashion.}
The core assumption underlying the methodology is that PLMs hold enough syntactic knowledge to be utilized for predicting parse trees by themselves.
Although the CPE-PLM framework has demonstrated that non-trivial trees resembling gold-standard annotations can be extracted from general PLMs even without fine-tuning on treebanks, it has been also reported that CPE-PLM's freedom from task-specific training comes at the cost of its performance inferior to that of unsupervised parsers.
Furthermore, as is the case of the foregoing unsupervised parsing literature, it has yet to be verified that CPE-PLM is also effective for languages other than English.

In this paper, we first attempt to narrow the performance gap between unsupervised parsers and CPE-PLM by introducing a novel method inspired by neural chart-based algorithms \cite{durrett-klein-2015-neural,stern-etal-2017-minimal,kitaev-klein-2018-constituency}.
In contrast to the top-down CPE-PLM method \cite{Kim2020Are}, which focuses on detecting the boundary of two subspans in a phrase relying only on the knowledge from the two words around the boundary, our chart-based method considers all components in a phrase to judge how plausible the phrase is.
Furthermore, we introduce a simple but effective ensemble technique that utilizes the pre-defined set of attention heads which are confirmed as being effective by their performance on the validation set.
We show that our chart-based method outperforms or is competitive to the top-down method and that the top-K ensemble plays a key role in boosting their performance. 

Second, the limitation of most previous studies for both unsupervised parsing and CPE-PLM is that they are heavily English-centric, leaving an open question whether they are universally applicable.
To investigate this problem, we test CPE-PLM on several other languages.
Specifically, we propose to introduce \textit{multilingual} PLMs \cite{NIPS2019_8928,conneau2019unsupervised} into CPE-PLM to grant the framework an ability to deal with multiple languages simultaneously.
We show that the CPE-PLM methods built upon multilingual PLMs are able to induce reasonable parses for sentences in nine languages in an integrated and language-agnostic manner, achieving figures superior or comparable to ones from neural PCFGs \cite{kim-etal-2019-compound,zhao2021empirical}.
In supplementary analyses, we provide intuitive explanations about the inner workings of our method. 
For instance, we confirm the existence of \textit{universal} attention heads which seem to be responsible for capturing syntactic information irrespective of the input language.
 
\begin{table}[t!]
\scriptsize
\setlength{\tabcolsep}{0.7em}
\begin{center}
\begin{tabular}{ccc}
\toprule
\bf Methodology  & \bf Unsupervised Parsing & \bf CPE-PLM \\ 
\midrule
\multirow{2}{*}{\bf Training data} & In-domain data & General corpora \\ 
& (e.g., raw text from PTB) & (e.g., Wikipedia) \\
\midrule
\multirow{2}{*}{\bf Architecture} & Task-oriented & \multirow{2}{*}{Transformer} \\ 
& (e.g., RNNG, PCFG) & \\
\midrule
\multirow{2}{*}{\bf Modeling} & $p(S,T)$ ($T$ is marginalized  & $p(S)$ ($T$ is not \\
& or implicitly modeled) & considered in modeling) \\
\bottomrule
\end{tabular}
\end{center}
\caption{Comparison between typical unsupervised parsing and constituency parse extraction from pre-trained language models (CPE-PLM).}
\label{table:table1}
\end{table}

\section{Background}

In this work, we focus on a variant of unsupervised constituency parsing, which we call \textsf{\small \textbf{C}onstituency \textbf{P}arse \textbf{E}xtraction from \textbf{P}re-trained \textbf{L}anguage \textbf{M}odels (\textbf{CPE-PLM})} \cite{Kim2020Are,wu-etal-2020-perturbed}.
We specify the characteristics of CPE-PLM in Table \ref{table:table1}, comparing them with those of general unsupervised parsing methods.

Typical unsupervised parsers consist of task-oriented architectures (e.g., RNNG \cite{kim-etal-2019-unsupervised} and PCFG \cite{kim-etal-2019-compound}) which are designed to model both a sentence $S$ and the corresponding tree $T$ (i.e., $p(S,T)$) and are trained with in-domain plain text.\footnote{As it is mostly infeasible to directly model the tree $T$ without supervision from gold annotations, unsupervised parsers usually make use of different approximation or marginalization techniques such as variational inference and sampling.} 
On the other hand, CPE-PLM simply employs off-the-shelf Transformer PLMs, which only model the probability of a sentence $p(S)$, as their core component and do not require additional training---the PLMs are frozen and no trainable component is augmented on top of them, meaning \textit{parameter-free}.
Instead, CPE-PLM methods take advantage of implicit syntactic knowledge residing in PLMs to reconstruct parses, by computing syntactic distances \cite{shen2018neural} between words in a sentence using features from the PLMs.
We describe their algorithmic details in Section \ref{sec: method}.
CPE-PLM's independence from training also makes it being distinct from syntactic probes for PLMs \cite{hewitt-manning-2019-structural,chi-etal-2020-finding} which demand training probing modules to investigate the latent knowledge of PLMs.


\section{Method} \label{sec: method}

Among various approaches that belong to CPE-PLM, we regard the top-down method proposed by \citet{Kim2020Are} as a starting point and aim to improve the method in several perspectives.

\begin{algorithm}[t!]
\small
\caption{Syntactic Distance to Binary Constituency Tree (from \citet{Kim2020Are})}
\label{alg:distance2tree} 
\begin{algorithmic}[1]
\State {$S=[w_1, w_2, \dots, w_n]$: Words in a sentence of length $n$.}
\State {$\mathbf{d}=[d_1,d_2,\dots,d_{n-1}]$: Vector, each of whose elements $d_i$ is the syntactic distance between $w_i$ and $w_{i+1}$.}
\Function{D2T}{$S$, $\mathbf{d}$}
    \If {$\mathbf{d} = []$}
        \State {node $\gets$ Leaf($S[0]$)}
    \Else
        \State {$i \gets \mathrm{arg}\max_i (\mathbf{d})$}
        \State {$\mathrm{child}_l$ $\gets$ \textsc{D2T}($S_{\leq i}, \mathbf{d}_{<i}$)}
        \State {$\mathrm{child}_r$ $\gets$ \textsc{D2T}($S_{>i}, \mathbf{d}_{>i}$)}
        \State $\mathrm{node} \gets$ Node($\mathrm{child}_l$, $\mathrm{child}_r$)
    \EndIf
    \State \Return $\mathrm{node}$
\EndFunction
\end{algorithmic}
\end{algorithm}

\subsection{Top-down CPE-PLM} \label{subsec: top-down CPE-PLM}

\citet{Kim2020Are} proposed a zero-shot version of top-down constituency parsing
\cite{shen-etal-2018-straight}, where a concept of \textit{syntactic distance} \cite{shen2018neural} plays a vital role.
Formally, given a sequence of words in a sentence $S=[w_1, w_2, \dots, w_n]$, the corresponding syntactic distance vector $\mathbf{d} = [d_1, d_2, \dots, d_{n-1}]$ is computed as follows (each $d_i$ is the syntactic distance between $w_i$ and $w_{i+1}$):
\begin{equation*}
    d_i = f(g(w_i), g(w_{i+1})),
\end{equation*}
where $f(\cdot,\cdot)$ and $g(\cdot)$ are a distance measure function and representation extractor function. 
For $g$, the authors suggest utilizing $G^v$ and $G^d$. 
Given $l$ as the number of layers in a PLM and $a$ as the number of attention heads per layer, $G^v$ refers to a set of functions $\{g_j^v|j=1,\dots,l\}$, each of which outputs the hidden representation of a given word on the $j^{th}$ layer of the PLM.
Similarly, $G^d$ is defined as $\{g_{(j,k)}^d|j=1,\dots,l, k=1,\dots,a+1\}$\footnote{Given $a$ attention heads on the $j^{th}$ layer, \citet{Kim2020Are} also consider the $(a+1)^{th}$ head that corresponds to the average of all attention distributions on the $j^{th}$ layer.}, each of whose elements computes the attention distribution of an input word by using the $k^{th}$ attention head on the $j^{th}$ layer of the PLM.
For the function $f$, there also exist two options, i.e., $F^v = \{\textsc{COS}, \textsc{L1}, \textsc{L2}\}$ and $F^d = \{\textsc{JSD}, \textsc{HEL}\}$, where COS, L1, L2, JSD, and HEL correspond to the Cosine, L1, and L2, Jensen-Shannon, and Hellinger distance respectively. 
Note that $F^v$ is only compatible with $G^v$ while $F^d$ is only with $G^d$.

Finally, given the input sentence $S$ and syntactic distance vector $\mathbf{d}$, Algorithm \ref{alg:distance2tree} is adopted to induce a complete (binary) constituency parse tree, recursively splitting $S$ in a top-down manner.

\begin{algorithm}[t!]
\small
\caption{Chart to Syntactic Distance}
\label{alg:tree2distance} 
\begin{algorithmic}[1]
\State {$n$: Length of an input sentence $S$.}
\State {$ C \in \mathbb{R}^{n \times n}$: Chart matrix whose elements are $s_{span}(i, j)$.}
\State {$ P \in \mathbb{R}^{n \times n}$: Matrix, whose $(i,j)^{th}$ element is the split point of the span $(i,j)$ of the sentence $S$.}
\State {$s$: Start position, initialized as $1$.}
\State {$e$: End position, initialized as $n$.}
\Function{C2D}{$C$, $P$, $s$, $e$}
    \If {$s = e$}
        \State {\Return [] (empty vector)}
    \Else
        \State {$v \gets C[s][e]$}
        \State {$p \gets P[s][e]$}
        \State {\Return [\textsc{C2D}($C$, $P$, $s$, $p$); $v$; \textsc{C2D}($C$, $P$, $p$+1, $e$)]}
        \State {($[\cdot;\cdot]$: vector concatenation)}
    \EndIf
\EndFunction
\end{algorithmic}
\end{algorithm}

\subsection{Chart-based CPE-PLM}

Although the top-down method has shown its effectiveness in extracting non-trivial phrase structures from PLMs, there still remains much room for improvement, considering that this method by nature operates in a greedy fashion rather than taking account of the probabilities of all possible subtrees.
In other words, the top-down CPE-PLM method only relies on the information obtained from the representations of two words to estimate the likelihood of the space between the two words becoming a target to be split.
To overcome this limitation, we propose a novel approach based on chart parsing which executes an exact inference to find the most probable parse while effectively considering all possibilities with dynamic programming.

Following the previous work on chart parsing \cite{stern-etal-2017-minimal,kitaev-klein-2018-constituency}, 
we assign a real-valued score $s_{tree}(T)$ for each tree candidate $T$, which decomposes as $s_{tree}(T) = \sum_{(i,j) \in T} s_{span}(i, j),$
where $s_{span}(i, j)$ is a score (or cost) for a constituent that is located between positions $i$ and $j$ ($1 \le i \le j \le n$) in a sentence.
Specifically, $s_{span}(i, j)$ is defined as follows:
\begin{equation*}
    \resizebox{.99\linewidth}{!}{
    $s_{span}(i, j) = 
    \begin{cases}
        s_{comp}(i,j) + \min_{i \le k < j}s_{split}(i, k, j) & \text{if} \ \ i < j \\
        0 & \text{if} \ \ i = j
    \end{cases}$}
\end{equation*}
where $s_{split}(i, k, j) = s_{span}(i, k) + s_{span}(k+1, j)$.
In other words, $s_{comp}(i,j)$ measures the validity or compositionality of the span $(i,j)$ itself while $s_{split}(i, k, j)$ indicates how plausible it is to divide the span $(i,j)$ into two subspans $(i,k)$ and $(k+1, j)$. 
We choose the most probable $k$ that brings us the minimum cost of  $s_{split}(i,k,j)$.
Note that each constituent is by definition evaluated with the scores of its children in addition to its own score.
Once $s_{comp}(\cdot, \cdot)$ is properly defined, it is straightforward to compute every $s_{span}(i,j)$ by utilizing the CKY algorithm \cite{cocke1969programming,kasami1966efficient,younger1967recognition}.
In the following subsections, we formulate two variants of $s_{comp}(\cdot, \cdot)$ in detail.

Finally, our parser outputs $\hat{T}$, the tree that requires the lowest score (cost) to build, as a prediction for the parse tree of the input sentence: $\hat{T} = \argmin_{T}s_{tree}(T)$.

\subsubsection{Pair Score Function} \label{subsubsec: pair score function}

The methodology introduced in the previous section abstracted over the choice of $s_{comp}(\cdot, \cdot)$; in what follows we propose two candidates for it.

First, we propose a pair score function $s_{p}(\cdot,\cdot)$ which is is defined as follows:
\begin{equation*}
    \resizebox{.99\linewidth}{!}{
    $s_{p}(i,j) := \binom{j-i+1}{2}^{-1} \sum_{(w_x,w_y)\in pair(i,j)} f(g(w_x), g(w_y)),$}
\end{equation*}
where $pair(i,j)$ returns a set consisting of all combinations of two words from a span $(i,j)$---e.g., $pair(1,3) = \{(w_1,w_2), (w_1,w_3), (w_2,w_3)\}$.
The intuition behind this formulation is that every pair of words in a constituent should have similar attention distributions so that the pair's embeddings become similar to each other in the subsequent layers of PLMs.
For $f(\cdot, \cdot)$ and $g(\cdot)$, we again take advantage of $F^d$ and $G^d$ specified in Section \ref{subsec: top-down CPE-PLM}.\footnote{This implies that we make only use of the attention distributions of PLMs,
as it is verified by the previous work \cite{Kim2020Are} and our preliminary experiments that attention distributions offer more useful signals in this setting.}

\subsubsection{Characteristic Score Function} \label{subsubsec: characteristic score function}

We also propose another candidate for $s_{comp}(\cdot,\cdot)$, namely a characteristic score function $s_{c}(\cdot,\cdot)$. 
Instead of measuring the similarities of all pairs of attention distributions, we pre-define $\boldsymbol{c}$ as the characteristic vector of a given constituent and evaluate the cost of each word in the constituent with regard to this value.
Although $\boldsymbol{c}$ can be realized in many ways, for simplicity, we here use the average of all the attention distributions of words in a constituent.
As a consequence, $s_{c}(i,j)$ is formalized as follows: 
\begin{equation*}
    s_{c}(i,j) := \frac{1}{j-i+1} \sum_{i\le x \le j} f(g(w_x), \boldsymbol{c}),
\end{equation*}
where $\boldsymbol{c} = \frac{1}{j-i+1} \sum_{i \le y \le j} g(w_y)$.

\subsection{Top-K Ensemble for CPE-PLM}

The part remaining ambiguous so far in clarifying CPE-PLM algorithms is about how to properly select the distance measure function $f$ and representation extractor function $g$ from the set of candidates, i.e., $F^d$ and $G^d$.
Basically, we can consider a typical case where we acquire the best combination of $f$ and $g$ using the validation set and apply it to the test set.
In addition, we introduce one more option, called \textit{top-K ensemble}, that enables us to integrate the knowledge from several attention heads.

Specifically, we first pick an arbitrary candidate for $f$, dubbed $\hat{f}$.\footnote{In practice, we test every element $f \in F^d$ exhaustively to select the best one as $F^d$ consists of only two elements.}
Then, we compute the parsing performance of every possible combination of $\hat{f}$ and $g \in G^d$ on the validation set and sort $G^d$ according to the performance of its elements.
After that, we simply choose the top $K$ elements from the sorted $G^d$ and allow all of them ($G^d_{topK}$) to participate in parsing instead of just leveraging the best single one.
At test time, given an input sentence, we predict $K$ separate trees using every element from $G^d_{topK}$, and then convert the trees into corresponding syntactic distance vectors (Algorithm \ref{alg:tree2distance}).
Finally, we compute the average of the syntactic distance vectors and translate this averaged vector into the final tree prediction (Algorithm \ref{alg:distance2tree}).

By introducing the top-K ensemble technique, it becomes possible to obtain a more accurate tree prediction while seamlessly combining diverse syntactic signals provided by different attention heads.

\begin{figure*}[t!]
\begin{center}
\includegraphics[width=0.99\linewidth]{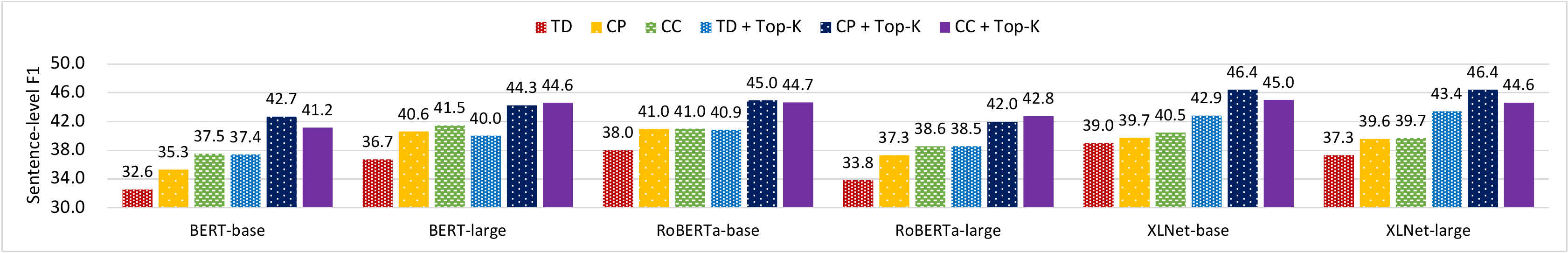}
\caption{
Performance of CPE-PLM methods on PTB.
Chart-based (CP and CC) approaches show superior figures in most cases compared to TD. 
The top-K ensemble also provides orthogonal improvements. Best viewed in color.} 
\label{fig:figure1}
\end{center}
\end{figure*}

\section{Experiments} \label{sec: experiments}

\subsection{General Configurations} \label{subsec: general settings}

To evaluate, we prepare the PTB \cite{marcus-etal-1993-building} dataset for English and the SPMRL \cite{seddah-etal-2013-overview} dataset for the eight other languages: Basque, French, German, Hebrew, Hungarian, Korean, Polish, and Swedish.
We use the standard split of each dataset, and the datasets are pre-processed following \citet{kim-etal-2019-compound} and \citet{zhao2021empirical}---removing punctuation marks.
We leverage the \textit{unlabelled sentence-level F1} (percentage) score  as a primary metric to evaluate the extent to which tree predictions resemble corresponding gold-standard trees.
The hyperparameter $K$, which determines the number of attention heads engaging in the top-K ensemble, is decided by grid search on some reasonable candidates ($\{5, 10, 20, 30\}$). 
We empirically found that $K$=20 is versatile across different settings.
From now on, we employ the abbreviations TD, CP, and CC to refer to the \textit{top-down} (Section \ref{subsec: top-down CPE-PLM}; \citet{Kim2020Are}), \textit{chart-pair} (Section \ref{subsubsec: pair score function}; our approach with $s_p$), and \textit{chart-characteristic} method (Section \ref{subsubsec: characteristic score function}; our method with $s_c$) respectively.

\begin{table}[t!]
\centering
\begin{center}
\setlength{\tabcolsep}{0.2em}
\extrarowheight=\aboverulesep
\addtolength{\extrarowheight}{\belowrulesep}
\aboverulesep=0pt
\belowrulesep=0pt
\resizebox{0.99\linewidth}{!}{%
\begin{tabular}{l aaa ccc aaa ccc}
\toprule
\bf Language & \multicolumn{3}{a}{\bf French} & \multicolumn{3}{c}{\bf German} & \multicolumn{3}{a}{\bf Korean} & \multicolumn{3}{c}{\bf Swedish} \\
\cmidrule(lr){2-4} \cmidrule(lr){5-7} \cmidrule(lr){8-10} \cmidrule(lr){11-13} 
\bf Method & \bf TD & \bf CP & \bf CC & \bf TD & \bf CP & \bf CC & \bf TD & \bf CP & \bf CC & \bf TD & \bf CP & \bf CC \\
\midrule
\textbf{Na\"ive baselines} & & & & & & & & & & & & \\
Random & \multicolumn{3}{a}{16.2} & \multicolumn{3}{c}{13.9} & \multicolumn{3}{a}{22.2} & \multicolumn{3}{c}{16.4} \\
Left-branching & \multicolumn{3}{a}{5.7} & \multicolumn{3}{c}{10.0} & \multicolumn{3}{a}{18.5} & \multicolumn{3}{c}{8.4} \\
Right-branching & \multicolumn{3}{a}{26.4} & \multicolumn{3}{c}{14.7} & \multicolumn{3}{a}{19.2} & \multicolumn{3}{c}{30.4} \\
\midrule
\textbf{PCFGs} & & & & & & & & & & & & \\
N-PCFG$^{\dagger}$ & \multicolumn{3}{a}{42.2} & \multicolumn{3}{c}{37.8} & \multicolumn{3}{a}{25.7} & \multicolumn{3}{c}{14.5} \\ 
C-PCFG$^{\dagger}$ & \multicolumn{3}{a}{40.5} & \multicolumn{3}{c}{37.3} & \multicolumn{3}{a}{27.7} & \multicolumn{3}{c}{23.7}\\
\midrule
\textbf{Mono. PLMs} & 41.4 & 42.4 & \textbf{42.8} & 38.4 & 39.6 & \textbf{39.7} & \textbf{51.1} & 47.3 & 47.4 & 35.6 & 38.4 & \textbf{38.9} \\
\bottomrule
\end{tabular}
}
\end{center}
\caption{CPE-PLM's performance on French, German, Korean, and Swedish. The best score for each language is in \textbf{bold}. $\dagger$: results from \citet{zhao2021empirical}.}
\label{table:table2}
\end{table}

\subsection{Experiments on Monolingual Settings} \label{subsec: experiments on monolingual settings}

We first assess CPE-PLM on the PTB dataset.
We apply our methods to three different categories of English PLMs---BERT \cite{devlin-etal-2019-bert}, RoBERTa \cite{liu2019roberta}, and XLNet \cite{yang2019xlnet}.\footnote{We prepare two variants for each PLM: (i) \textit{X-base} consists of 12 layers, 12 attention heads, and 768 hidden dimensions. (ii) \textit{X-large} has 24 layers, 16 heads, and 1024 dimensions.}
We also test the effect of the top-K ensemble by combining it with each of the CPE-PLM method.
In Figure \ref{fig:figure1}, we confirm that our chart-based methods mostly outperform the top-down approach, showing an improvement of up to nearly five points (RoBERTa-large: 33.8 $\rightarrow$ 38.6).
Moreover, we reveal that the top-K ensemble provides significant improvements on parsing performance in an orthogonal manner, regardless of the accompanying method.
This result implies that the cues that can contribute to inducing parse trees may be distributed across different parts of PLMs, rather than concentrated on a specific attention head.
We attain the 46.4 F1 score by combining XLNet, the CP method, and the top-K ensemble, which is six points higher than the best performance (40.1) reported in the previous work \cite{Kim2020Are}.

Next, we evaluate CPE-PLM methods on French, German, Korean, and Swedish, for which language-specific BERT variants are available.\footnote{The PLMs used per language are listed in Appendix.}
As baselines, we prepare three na\"ive methods---random and left/right-branching trees---in addition to N(eural)-PCFG and C(ompound)-PCFG \cite{kim-etal-2019-compound}, which are representative unsupervised parsers.\footnote{We rely on \citet{zhao2021empirical} to report PCFG models' performance. Note that the authors assume they have access to parses in the PTB dev set to select the best hyperparameters following \citet{kim-etal-2019-compound}. This condition is exactly the same for our English models and cross-lingual transfer settings in Table \ref{table:table3}. Meanwhile, we additionally employ the SPMRL validation set in Table \ref{table:table2} and the Multi-ling. part of Table \ref{table:table3}, which can give undesirable extra gains to CPE-PLM.}
For CPE-PLM, we consider both the top-down and chart methods, all of which are combined with the top-K ensemble.
In Table \ref{table:table2}, the CPE-PLM methods demonstrate performance comparable (French / German) or superior (Korean / Swedish) to that of the strong baselines.
In particular, CPE-PLM shows much better performance in Korean and Swedish, where PCFGs failed to obtain meaningful results.
We conjecture this discrepancy in part comes from the availability of subword-level features, to which PLMs have access while PCFGs do not, considering that the SPMRL dataset is originally constructed for testing morphologically-rich languages.
Meanwhile, CP and CC outperform TD on 3 out of 4 languages, albeit the gaps are relatively small compared to the English case.
This outcome leads to two implications: (i) the effectiveness of a CPE-PLM method depends on the language where it is applied, and (ii) our top-K ensemble is broadly helpful for all the parsing methods, reducing the gap between their performance.

\begin{table*}[t!]
\begin{center}
\setlength{\tabcolsep}{0.2em}
\extrarowheight=\aboverulesep
\addtolength{\extrarowheight}{\belowrulesep}
\aboverulesep=0pt
\belowrulesep=0pt
\resizebox{0.99\linewidth}{!}{%
\begin{tabular}{l aaa ccc aaa ccc aaa ccc aaa ccc aaa}
\toprule
\bf Language & \multicolumn{3}{a}{\bf English} & \multicolumn{3}{c}{\bf Basque} & \multicolumn{3}{a}{\bf French} & \multicolumn{3}{c}{\bf German} & \multicolumn{3}{a}{\bf Hebrew} & \multicolumn{3}{c}{\bf Hungarian} & \multicolumn{3}{a}{\bf Korean} & \multicolumn{3}{c}{\bf Polish} & \multicolumn{3}{a}{\bf Swedish} \\ 
\cmidrule(lr){2-4} \cmidrule(lr){5-7} \cmidrule(lr){8-10} \cmidrule(lr){11-13} \cmidrule(lr){14-16} \cmidrule(lr){17-19} \cmidrule(lr){20-22} \cmidrule(lr){23-25} \cmidrule(lr){26-28}
\bf Method & \bf TD & \bf CP & \bf CC & \bf TD & \bf CP & \bf CC & \bf TD & \bf CP & \bf CC & \bf TD & \bf CP & \bf CC & \bf TD & \bf CP & \bf CC & \bf TD & \bf CP & \bf CC & \bf TD & \bf CP & \bf CC & \bf TD & \bf CP & \bf CC & \bf TD & \bf CP & \bf CC \\
\midrule
\textbf{PCFGs} & & & & & & & & & & & & & & & & & & & & & & & & & & & \\ 
N-PCFG$^{\dagger}$ & \multicolumn{3}{a}{50.8} & \multicolumn{3}{c}{30.2} & \multicolumn{3}{a}{42.2} & \multicolumn{3}{c}{37.8} & \multicolumn{3}{a}{41.0} & \multicolumn{3}{c}{37.9} & \multicolumn{3}{a}{25.7} & \multicolumn{3}{c}{31.7} & \multicolumn{3}{a}{14.5} \\ 
C-PCFG$^{\dagger}$ & \multicolumn{3}{a}{\underline{55.7}} & \multicolumn{3}{c}{27.9} & \multicolumn{3}{a}{40.5} & \multicolumn{3}{c}{37.3} & \multicolumn{3}{a}{39.2} & \multicolumn{3}{c}{38.3} & \multicolumn{3}{a}{27.7} & \multicolumn{3}{c}{32.4} & \multicolumn{3}{a}{23.7} \\
\midrule
\textbf{Mono-ling.} & 42.9 & 46.4 & 45.0 & & & & 41.4 & 42.4 & 42.8 & 38.4 & 39.6 & 39.7 & & & & & & & 51.1 & 47.3 & 47.4 & & & & 35.6 & 38.4 & 38.9 \\ 
\midrule
\textbf{Multi-ling.}  & & & & & & & & & & & & & & & & & & & & & & & & & & & \\ 
M-BERT & 40.3 & \textbf{45.0} & 44.8 & 40.0 & 40.9 & \textbf{41.6} & 42.9 & 44.6 & \textbf{45.6} & 39.3 & \textbf{40.6} & 40.3 & 42.3 & \textbf{42.5} & 42.0 & 38.2 & 39.1 & \textbf{40.4} & \textbf{52.1} & 50.9 & 49.8 & 41.6 & \textbf{43.0} & 42.9 & 37.3 & 37.4 & \textbf{39.3} \\ 
XLM & 44.2 & \textbf{47.7} & 46.2 & 43.3 & 44.1 & \textbf{44.7} & 43.7 & \textbf{46.0} & \textbf{46.0} & 39.2 & \textbf{41.3} & 40.4 & 43.9 & 44.2 & \textbf{44.3} & 40.8 & \textbf{42.3} & 42.2 & \textbf{43.0} & 41.6 & 41.0 & 44.3 & \textbf{44.9} & 44.4 & 39.0 & 40.1 & \textbf{40.7} \\
XLM-R & 45.5 & 46.7 & \textbf{47.0} & 43.7 & 43.8 & \textbf{45.1} & \textbf{45.8} & 44.2 & 45.5 & 41.4 & \underline{\textbf{42.2}} & 41.6 & 45.0 & 43.2 & \textbf{45.3} & 42.4 & \underline{\textbf{44.0}} & 43.4 & \underline{\textbf{55.9}} & 55.7 & 54.3 & 43.1 & 43.7 & \textbf{44.6} & 39.5 & 40.6 & \underline{\textbf{41.5}} \\
XLM-R-L & 41.7 & 44.6 & \textbf{45.1} & 44.3 & 44.1 & \underline{\textbf{45.2}} & 39.5 & 42.4 & \textbf{42.9} & 38.9 & \textbf{41.0} & 40.7 & 43.7 & 44.1 & \underline{\textbf{46.3}} & 39.8 & \textbf{41.5} & 41.3 & 51.8 & \textbf{52.6} & 51.8 & 41.7 & 43.5 & \textbf{44.5} & 36.2 & 38.6 & \textbf{39.4} \\ 
\midrule
\textbf{Cross-ling.} & & & & & & & & & & & & & & & & & & & & & & & & & & & \\
M-BERT & & & & 39.8 & 39.8 & \textbf{41.1} & 42.2 & 44.6 & \textbf{45.5} & 37.7 & \textbf{40.3} & 39.3 & 39.7 & \textbf{42.8} & \textbf{42.8} & 36.2 & \textbf{39.4} & 38.0 & \textbf{48.9} & 47.0 & 45.7 & 39.8 & 41.9 & \textbf{42.3} & 36.0 & \textbf{39.1} & 38.7 \\
\hspace{2mm}($+$,$-$) & & & & \textcolor{blue}{(-0.2)} & \textcolor{blue}{(-1.1)} & \textcolor{blue}{(-0.5)} & \textcolor{blue}{(-0.7)} & (0.0) & \textcolor{blue}{(-0.1)} & \textcolor{blue}{(-1.6)} & \textcolor{blue}{(-0.3)} & \textcolor{blue}{(-1.0)} & \textcolor{blue}{(-2.6)} & \textcolor{red}{(+0.3)} & \textcolor{red}{(+0.8)} & \textcolor{blue}{(-2.0)} & \textcolor{red}{(+0.3)} & \textcolor{blue}{(-2.4)} & \textcolor{blue}{(-3.2)} & \textcolor{blue}{(-3.9)} & \textcolor{blue}{(-4.1)} & \textcolor{blue}{(-1.8)} & \textcolor{blue}{(-1.1)} & \textcolor{blue}{(-0.6)} & \textcolor{blue}{(-1.3)} & \textcolor{red}{(+1.7)} & \textcolor{blue}{(-0.6)} \\
\hdashline
XLM & & & & 40.6 & 41.2 & \textbf{42.1} & 44.2 & \underline{\textbf{46.3}} & 46.1 & 38.9 & \textbf{41.5} & 40.3 & 42.4 & \textbf{45.8} & 43.9 & 38.0 & \textbf{42.2} & 40.7 & \textbf{40.1} & 39.5 & 38.4 & 42.2 & \textbf{44.5} & 44.4 & 38.2 & \textbf{40.9} & \textbf{40.9} \\ 
\hspace{2mm}($+$,$-$) & & & & \textcolor{blue}{(-2.7)} & \textcolor{blue}{(-2.9)} & \textcolor{blue}{(-2.6)} & \textcolor{red}{(+0.5)} & \textcolor{red}{(+0.3)} & \textcolor{red}{(+0.1)} & \textcolor{blue}{(-0.3)} & \textcolor{red}{(+0.2)} & \textcolor{blue}{(-0.1)} & \textcolor{blue}{(-1.5)} & \textcolor{red}{(+1.6)} & \textcolor{blue}{(-0.4)} & \textcolor{blue}{(-2.8)} & \textcolor{blue}{(-0.1)} & \textcolor{blue}{(-1.5)} & \textcolor{blue}{(-2.9)} & \textcolor{blue}{(-2.1)} & \textcolor{blue}{(-2.6)} & \textcolor{blue}{(-2.1)} & \textcolor{blue}{(-0.4)} & (0.0) & \textcolor{blue}{(-0.8)} & \textcolor{red}{(+0.8)} & \textcolor{red}{(+0.2)} \\
\hdashline
XLM-R & & & & 43.4 & 42.1 & \textbf{43.7} & 45.4 & 45.1 & \textbf{46.2} & 41.5 & \underline{\textbf{42.2}} & 41.5 & 45.5 & 45.2 & \underline{\textbf{46.3}} & 41.3 & \textbf{43.4} & 41.9 & \textbf{52.6} & 49.6 & 48.9 & 44.3 & \underline{\textbf{45.4}} & 44.8 & 40.4 & 41.0 & \textbf{41.4} \\ 
\hspace{2mm}($+$,$-$) & & & & \textcolor{blue}{(-0.3)} & \textcolor{blue}{(-1.7)} & \textcolor{blue}{(-1.4)} & \textcolor{blue}{(-0.4)} & \textcolor{red}{(+0.9)} & \textcolor{red}{(+0.7)} & \textcolor{red}{(+0.1)} & (0.0) & \textcolor{blue}{(-0.1)} & \textcolor{red}{(+0.5)} & \textcolor{red}{(+2.0)} & \textcolor{red}{(+1.0)} & \textcolor{blue}{(-1.1)} & \textcolor{blue}{(-0.6)} & \textcolor{blue}{(-1.5)} & \textcolor{blue}{(-3.3)} & \textcolor{blue}{(-6.1)} & \textcolor{blue}{(-5.4)} & \textcolor{red}{(+1.2)} & \textcolor{red}{(+1.7)} & \textcolor{red}{(+0.2)} & \textcolor{red}{(+0.9)} & \textcolor{red}{(+0.4)} & \textcolor{blue}{(-0.1)} \\
\hdashline
XLM-R-L & & & & \textbf{43.9} & 42.6 & 43.6 & 39.4 & 42.3 & \textbf{43.2} & 38.6 & \textbf{40.6} & \textbf{40.6} & 42.8 & 44.7 & \textbf{45.4} & 38.6 & 39.9 & \textbf{40.7} & \textbf{51.6} & 51.3 & 50.5 & 42.6 & 44.9 & \textbf{45.1} & 37.1 & 39.6 & \textbf{40.0} \\
\hspace{2mm}($+$,$-$) & & & & \textcolor{blue}{(-0.4)} & \textcolor{blue}{(-1.5)}	& \textcolor{blue}{(-1.6)} & \textcolor{blue}{(-0.1)} & \textcolor{blue}{(-0.1)} & \textcolor{red}{(+0.3)} & \textcolor{blue}{(-0.3)} & \textcolor{blue}{(-0.4)} & \textcolor{blue}{(-0.1)} & \textcolor{blue}{(-0.9)} & \textcolor{red}{(+0.6)} & \textcolor{blue}{(-0.9)} & \textcolor{blue}{(-1.2)} & \textcolor{blue}{(-1.6)} & \textcolor{blue}{(-0.6)} & \textcolor{blue}{(-0.2)} & \textcolor{blue}{(-1.3)} & \textcolor{blue}{(-1.3)} & \textcolor{red}{(+0.9)} & \textcolor{red}{(+1.4)} & \textcolor{red}{(+0.6)} & \textcolor{red}{(+0.9)} & \textcolor{red}{(+1.0)} & \textcolor{red}{(+0.6)} \\
\bottomrule
\end{tabular}
}
\end{center}
\caption{Performance of CPE-PLM on 9 languages.
\textbf{Mono-ling.}: CPE-PLM's performance in monolingual settings.
\textbf{Multi-ling.}: the results when combined with multilingual PLMs.
\textbf{Cross-ling.}: the performance when relying on cross-lingual transfer, in addition to the relative losses or gains ($+$,$-$) compared to the original results.
The best score per PLM is in \textbf{bold} while the best for each language is \underline{underlined}.
$\dagger$: results from \citet{zhao2021empirical}.}
\label{table:table3}
\end{table*}

\subsection{Experiments on Multilingual Settings} \label{subsec: experiments on multilingual settings}

Theoretically, multilingual PLMs have a potential to be a core asset for CPE-PLM, given that they are able to deal with sentences from over a hundred languages simultaneously.
However, it has not yet been investigated whether they can play a role as expected.
To shed light on this issue, we conduct experiments with the CPE-PLM methods built upon multilingual PLMs.
We apply four multilingual PLMs to nine languages in total.
We use a multilingual version of the BERT-base model (M-BERT, \citet{devlin-etal-2019-bert}), the XLM model trained on 100 languages (XLM, \citet{NIPS2019_8928}), XLM-R, and XLM-R-L(arge) \cite{conneau2019unsupervised}.
For baselines, we only consider PCFGs as we verified in Section \ref{subsec: experiments on monolingual settings} that they can subsume na\"ive baselines.
We also list CPE-PLM's performance in monolingual settings for reference.
Again, we utilize the TD, CP, and CC methods combined with the top-K ensemble.

In the \textbf{Multi-ling.} section of Table \ref{table:table3}, we report CPE-PLM's performance with multilingual PLMs when the best attention heads are separately selected for each language, relying on the validation sets of respective languages.
We observe that the CPE-PLM framework works pretty well across languages when it is built upon multilingual PLMs, outperforming PCFGs except for English.
Surprisingly, we discover that for every language we consider, there exists at least one multilingual PLM that outperforms its monolingual counterpart.
For instance, we achieve the F1 score of 47.7 in English with the XLM model, which is higher than all the scores we achieved for English in monolingual settings.
In conclusion, we confirm that multilingual PLMs can serve as a core component for an integrated CPE-PLM framework that processes different languages simultaneously.
Regarding the effect of parsing strategies, we identify that CP and CC generally outperform TD, and that the only exception occurs in Korean.
We assume this is related to the linguistic properties of target languages, but we leave a thorough analysis on this as future work.

Next, we evaluate CPE-PLM in a harsher condition where the validation set is given only for English.
Concretely, we attempt to parse sentences in eight other languages with the CPE-PLM methods optimized for English (i.e., the attention heads are chosen based on the PTB validation set), performing \textit{zero-shot cross-lingual transfer} from English to others.
Note that this constraint facilitates revealing the true value of CPE-PLM by answering the following research question: \textit{given no access to parsers or gold-standard trees in target languages at all, can we induce non-trivial parse trees by solely relying on the knowledge residing in PLMs?}

\begin{table*}[t!]
\begin{center}
\setlength{\tabcolsep}{0.2em}
\extrarowheight=\aboverulesep
\addtolength{\extrarowheight}{\belowrulesep}
\aboverulesep=0pt
\belowrulesep=0pt
\resizebox{0.99\linewidth}{!}{%
\begin{tabular}{l aaa ccc aaa ccc aaa ccc aaa ccc aaa}
\toprule
\bf Language & \multicolumn{3}{a}{\bf English} & \multicolumn{3}{c}{\bf Basque} & \multicolumn{3}{a}{\bf French} & \multicolumn{3}{c}{\bf German} & \multicolumn{3}{a}{\bf Hebrew} & \multicolumn{3}{c}{\bf Hungarian} & \multicolumn{3}{a}{\bf Korean} & \multicolumn{3}{c}{\bf Polish} & \multicolumn{3}{a}{\bf Swedish} \\ 
\cmidrule(lr){2-4} \cmidrule(lr){5-7} \cmidrule(lr){8-10} \cmidrule(lr){11-13} \cmidrule(lr){14-16} \cmidrule(lr){17-19} \cmidrule(lr){20-22} \cmidrule(lr){23-25} \cmidrule(lr){26-28}
\bf Method & \bf TD & \bf CP & \bf CC & \bf TD & \bf CP & \bf CC & \bf TD & \bf CP & \bf CC & \bf TD & \bf CP & \bf CC & \bf TD & \bf CP & \bf CC & \bf TD & \bf CP & \bf CC & \bf TD & \bf CP & \bf CC & \bf TD & \bf CP & \bf CC & \bf TD & \bf CP & \bf CC \\
\midrule
\textbf{Pre-training data} & & & & & & & & & & & & & & & & & & & & & & & & & & & \\
Tokens (M)$^{\dagger}$  & \multicolumn{3}{a}{300.8} & \multicolumn{3}{c}{2.0} & \multicolumn{3}{a}{56.8} & \multicolumn{3}{c}{66.6} & \multicolumn{3}{a}{31.6} & \multicolumn{3}{c}{58.4} & \multicolumn{3}{a}{54.2} & \multicolumn{3}{c}{44.6} & \multicolumn{3}{a}{12.1} \\
Size (GiB)$^{\dagger}$ & \multicolumn{3}{a}{55608} & \multicolumn{3}{c}{270} & \multicolumn{3}{a}{9780} & \multicolumn{3}{c}{10297} & \multicolumn{3}{a}{3399} & \multicolumn{3}{c}{7807} & \multicolumn{3}{a}{5644} & \multicolumn{3}{c}{6490} & \multicolumn{3}{a}{77} \\
\midrule
\textbf{Val. \& Test data} & & & & & & & & & & & & & & & & & & & & & & & & & & & \\
Size (Validation) & \multicolumn{3}{a}{1700} & \multicolumn{3}{c}{948} & \multicolumn{3}{a}{1235} & \multicolumn{3}{c}{5000} & \multicolumn{3}{a}{500} & \multicolumn{3}{c}{1051} & \multicolumn{3}{a}{2066} & \multicolumn{3}{c}{821} & \multicolumn{3}{a}{494} \\
Size (Test) & \multicolumn{3}{a}{2416} & \multicolumn{3}{c}{946} & \multicolumn{3}{a}{2540} & \multicolumn{3}{c}{4999} & \multicolumn{3}{a}{716} & \multicolumn{3}{c}{1009} & \multicolumn{3}{a}{2287} & \multicolumn{3}{c}{822} & \multicolumn{3}{a}{666} \\
\midrule
\bf XLM-R & 45.5 & 46.7 & \textbf{47.0} & 43.7 & 43.8 & \textbf{45.1} & \textbf{45.8} & 44.2 & 45.5 & 41.4 & \textbf{42.2} & 41.6 & 45.0 & 43.2 & \textbf{45.3} & 42.4 & \textbf{44.0} & 43.4 & \textbf{55.9} & 55.7 & 54.3 & 43.1 & 43.7 & \textbf{44.6} & 39.5 & 40.6 & \textbf{41.5} \\
\hspace{2mm} Cross-lingual & & & & 43.4 & 42.1 & \textbf{43.7} & 45.4 & 45.1 & \textbf{46.2} & 41.5 & \textbf{42.2} & 41.5 & 45.5 & 45.2 & \textbf{46.3} & 41.3 & \textbf{43.4} & 41.9 & \textbf{52.6} & 49.6 & 48.9 & 44.3 & \textbf{45.4} & 44.8 & 40.4 & 41.0 & \textbf{41.4} \\
\hspace{4mm} ($+$,$-$) & & & & \textcolor{blue}{(-0.3)} & \textcolor{blue}{(-1.7)} & \textcolor{blue}{(-1.4)} & \textcolor{blue}{(-0.4)} & \textcolor{red}{(+0.9)} & \textcolor{red}{(+0.7)} & \textcolor{red}{(+0.1)} & (0.0) & \textcolor{blue}{(-0.1)} & \textcolor{red}{(+0.5)} & \textcolor{red}{(+2.0)} & \textcolor{red}{(+1.0)} & \textcolor{blue}{(-1.1)} & \textcolor{blue}{(-0.6)} & \textcolor{blue}{(-1.5)} & \textcolor{blue}{(-3.3)} & \textcolor{blue}{(-6.1)} & \textcolor{blue}{(-5.4)} & \textcolor{red}{(+1.2)} & \textcolor{red}{(+1.7)} & \textcolor{red}{(+0.2)} & \textcolor{red}{(+0.9)} & \textcolor{red}{(+0.4)} & \textcolor{blue}{(-0.1)} \\
\bottomrule
\end{tabular}
}
\end{center}
\caption{Factor correlation analysis. The first section describes the statistics of the data utilized for training XLM-R. The second section displays the characteristics of the validation and test sets.  $\dagger$: from \citet{conneau2019unsupervised}.}
\label{table:table4}
\end{table*}

In the \textbf{Cross-ling.} section, we present the performance of cross-lingual transfer and relative performance losses or gains ($+$,$-$) compared against the language-specific optimization (\textbf{Multi-ling.}).
To our surprise, we reveal that the cross-lingual transfer leads to negligible losses or even small gains in most cases.
This is also in line with the reports from related work (\citet{pires-etal-2019-multilingual,Cao2020Multilingual}; \textit{i.a.}) that multilingual PLMs are effective in cross/multi-lingual NLP tasks.
Our finding implies that there exist \textit{universal} attention heads that are sensitive to the phrase structures of sentences irrespective of the input language.
We seek to analyze this phenomenon in detail in the following section.

\section{Analysis} \label{sec: analysis}

We present several analyses that enrich our understanding about CPE-PLM.
We employ XLM-R as a backbone and the CC method as a parsing scheme.

\subsection{Factor Correlation Analysis} \label{subsec: factor analysis}

First, we attend to two factors that may affect CPE-PLM's performance: (i) the amount of the data consumed to train a PLM, and (ii) the number of sentences in the validation and test sets.
In Table \ref{table:table4}, we do not notice a clear relationship between the amount of pre-training data and performance.
We conjecture this result is rooted in the sampling technique exploited when pre-training XLM-R.
Specifically, the technique readjusts the probability of sampling a sentence from each language, increasing the number of tokens sampled from low-resource languages while mitigating the bias towards high-resource languages \cite{conneau2019unsupervised}. 
On the other hand, we discover that the languages for which the size of the validation sets are relatively small (i.e., Hebrew, Polish, and Swedish) tend to benefit from cross-lingual transfer, implying that the insufficient number of examples in the validation set might cause some noise or lead to the suboptimal in the selection of attention heads.

\begin{figure}[t!]
\begin{center}
\includegraphics[width=0.85\linewidth]{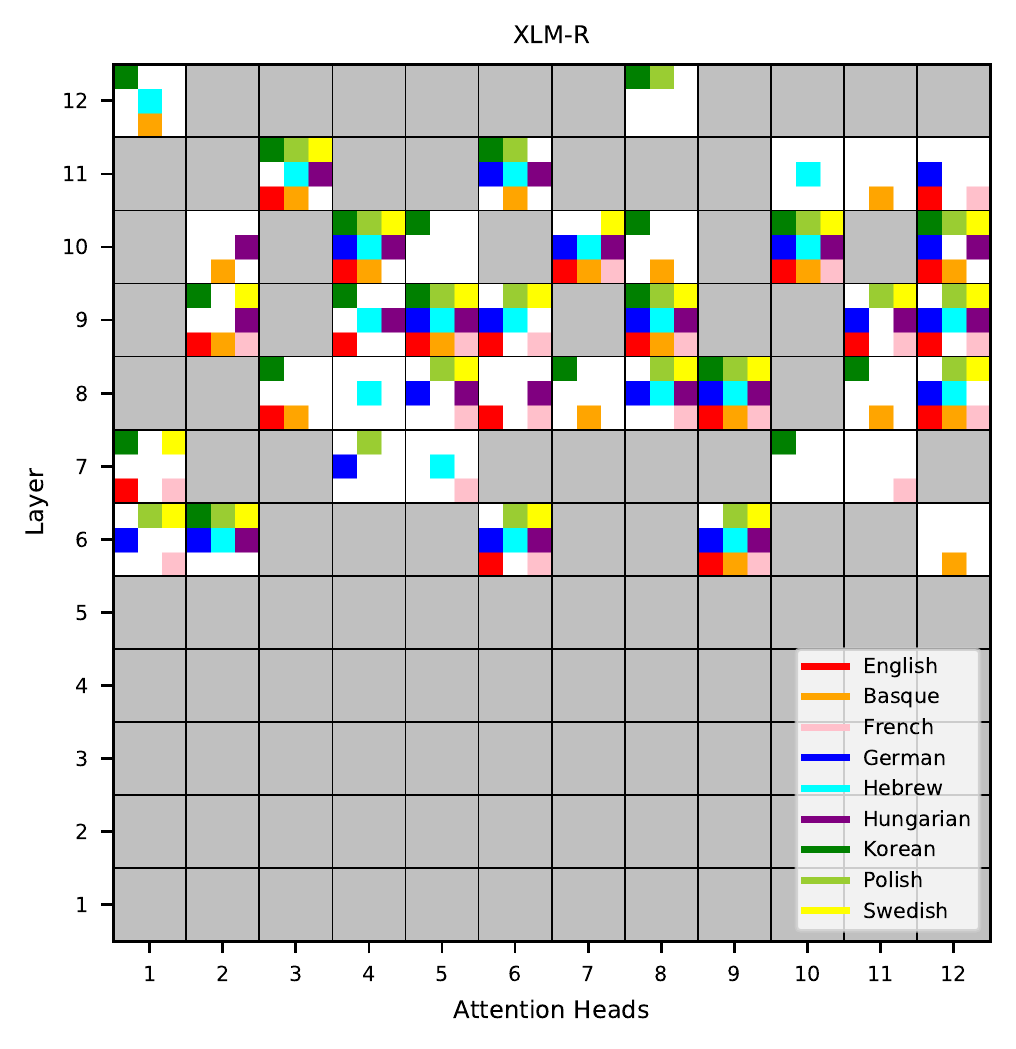}
\caption{Visualization of the sets of the top 20 attention heads (in XLM-R) for 9 languages. Each cell is filled with the color assigned for a language if the corresponding head is responsible for parsing the language.} 
\label{fig:figure2}
\end{center}
\end{figure}

\subsection{Visualization of Attention Heads}

We revealed in Section \ref{subsec: experiments on multilingual settings} that CPE-PLM's performance for most languages does not suffer much from cross-lingual transfer, suggesting that there would exist significant overlaps among the sets of the attention heads selected for respective languages.
To verify our hypothesis, in Figure \ref{fig:figure2}, we visualize the language-specific sets of the top 20 heads existing in XLM-R.
We observe that the heads effective for CPE-PLM are distributed over the middle-to-upper (6-12) layers of XLM-R, implying that phrase-level information is pervasive in the upper layers rather than the lower ones.
In addition, we discover that most of the heads detected as sensitive to syntax respond to multiple languages simultaneously and that there exist a few heads proven to be important for dealing with all the nine languages we consider.
Our finding of the existence of such \textit{universal} attention heads explains why CPE-PLM is robust to cross-lingual transfer in multilingual settings, in addition to providing a partial clue on why multilingual PLMs are excel at cross-lingual transfer as reported in previous work (\citet{pires-etal-2019-multilingual,Cao2020Multilingual}; \textit{inter alia}).

\begin{figure}[t!]
\begin{center}
\includegraphics[width=\linewidth]{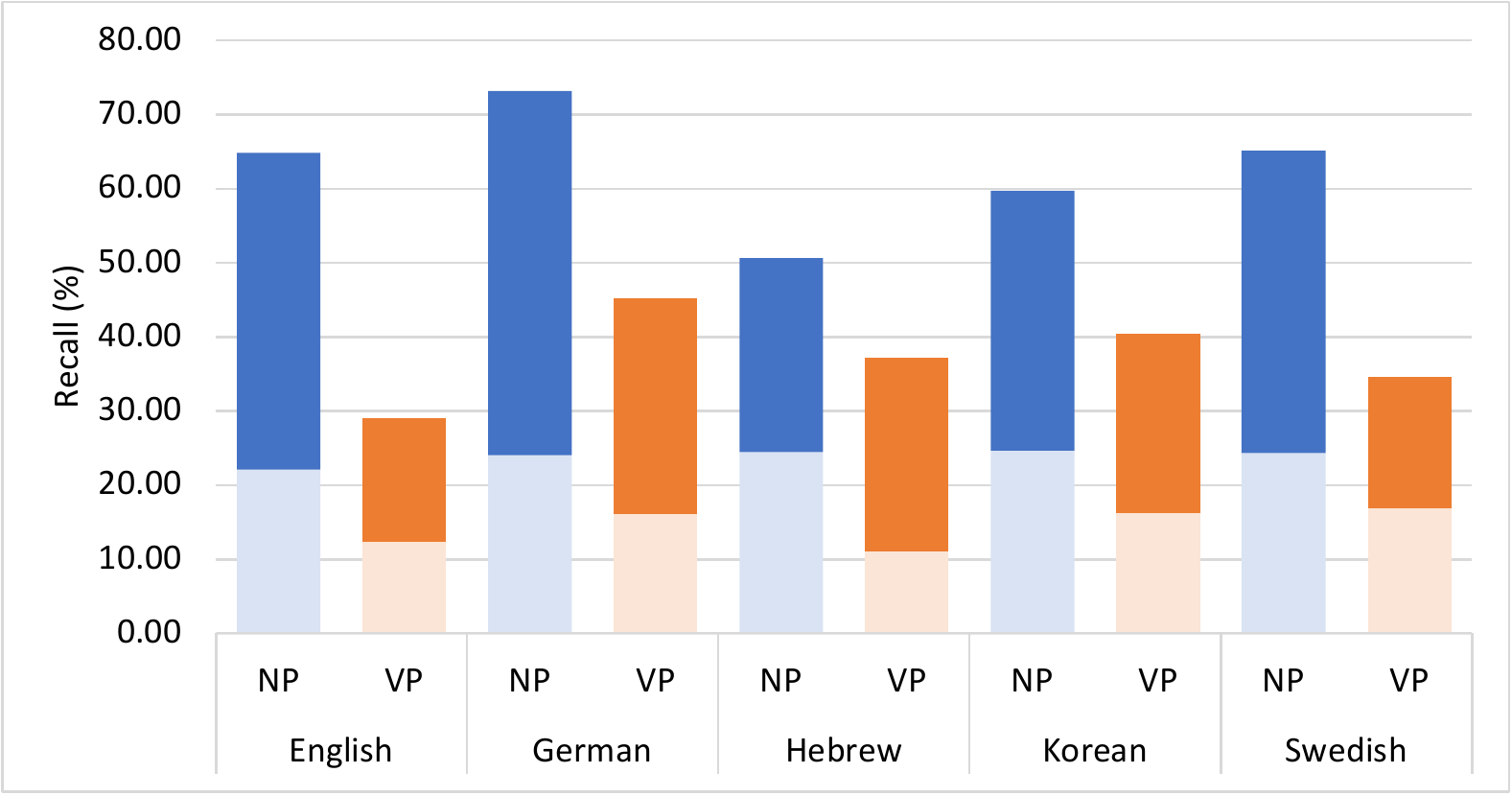}
\caption{Recall scores on gold-standard NPs and VPs.
The light bars indicate the random baseline's performance while the dark ones show that of the CC method.} 
\label{fig:figure3}
\end{center}
\end{figure}

\subsection{Recall Scores on Noun and Verb Phrases}

To assess CPE-PLM's performance in a more fine-grained manner and probe the extent to which it detects the core components of sentences, we present its recall scores on gold-standard noun and verb phrases.
We only target the languages whose gold-standard trees contain proper tags in their test sets.
In Figure \ref{fig:figure3}, we confirm that compared to the random baseline, CPE-PLM has a decent ability to identify noun phrases, succeeding in retrieving more than half of NPs for every language.
On the contrary, CPE-PLM seems relatively weak in recognizing VPs which are generally longer and more complex than NPs.
This implies that CPE-PLM might struggle with grasping the whole structure of sentences (e.g., VPs), although it successfully perceives small phrasal components (e.g., NPs).

\section{Limitations and Future Work}

We here mention a few limitations of our approach and propose avenues for future work.
First, analogous to several unsupervised parsers \cite{shi2020role} including PCFGs \cite{zhao2021empirical}, the current form of our method relies on a few gold-standard annotations from the validation set to determine the best hyperparameters (i.e., the best choice for attention head selection).
This dependency makes it hard to say that our approach is entirely unsupervised, although it steps aside from a typical way of learning parsers with supervision.
A next, promising yet challenging, step will be therefore to develop a remedy that enables our method to be free from the annotations, similar to \citet{li-etal-2020-heads}.
Note that our cross-lingual transfer experiments also shed some light on how to relieve such dependency.

While we have shown that CPE-PLM can be superior or comparable to PCFGs and that it can function as an effective tool for analyzing PLMs, its performance still falls short of expectations in terms of whether it can practically substitute standard parsers, similar to the case of unsupervised parsers.
To improve its performance, we have a plan as future work to design an ensemble method that gathers information from heterogeneous PLMs.

Finally, chart parsing algorithm, whose time complexity is $O(n^3)$, is inherently much more expensive than other efficient parsing strategies such as top-down parsing. 
Therefore, when we need to decide which parsing algorithm to employ, we should keep in mind the trade-off between accuracy and efficiency.

\section{Related work} \label{sec: related work}

Pre-trained language models (PLMs) now lie at the heart of many studies in the literature. 
Following the trend, much effort has been made to develop English PLMs (\citet{devlin-etal-2019-bert,liu2019roberta,radford2019language,yang2019xlnet}, \textit{inter alia}), to construct non-English PLMs (\citet{martin2019camembert}; \textit{i.a.}), and to train multilingual variants \cite{NIPS2019_8928,conneau2019unsupervised}.
We have explored the potential use of these PLMs as parsers.
The trees induced by our method can also be leveraged as a tool for probing PLMs, similar to recent work that attempt to explore the knowledge in PLMs \cite{clark-etal-2019-bert,jawahar-etal-2019-bert}.
In particular, \citet{chi-etal-2020-finding} extended \citet{hewitt-manning-2019-structural} to multilingual settings, analogous to our work. 
Still, it is different from ours in that it requires explicit supervision and devotes itself to dependency grammar.

In this study, we have named a line of work \cite{rosa2019inducing,wu-etal-2020-perturbed,Kim2020Are} that extracts trees directly from PLMs as CPE-PLM. 
We have extended its application to multilingual scenarios as well as improving its performance.
Notably, \citet{marecek-rosa-2019-balustrades} developed an approach similar to ours, but they focused on Transformers trained for machine translation rather than language models.
Work on neural unsupervised parsing (\citet{shen2018neural,shen2019ordered,kim-etal-2019-compound,shi2020role}, \textit{inter alia}) also seeks to generate parse trees without supervision from gold-standard trees.
It is worth noting that some work such as \citet{kann-etal-2019-neural} and \citet{zhao2021empirical} attempt to evaluate unsupervised parsers in multilingual settings, akin to our work.
The difference between ours and theirs is that our method does not require training a parser for each language, instead relying on off-the-shelf PLMs.

\section{Conclusion} \label{sec: conclusion}

In this work, we study \textsf{\small \textbf{C}onstituency \textbf{P}arse \textbf{E}xtraction from \textbf{P}re-trained \textbf{L}anguage \textbf{M}odels (\textbf{CPE-PLM})}, a novel paradigm of inducing parses directly from PLMs. 
We introduce a chart-based method and top-K ensemble for improving performance, and extend the range of application of the paradigm to different languages by applying multilingual PLMs.
We hope our work can function as the foundation for future research on (i) unsupervised constituency parsing for under-represented languages and (ii) probing the inner workings of multilingual PLMs.

\section*{Acknowledgements}

This work was supported by Institute of Information \& communications Technology Planning \& Evaluation (IITP) grant funded by the Korea government(MSIT) (No.2020-0-01373, Artificial Intelligence Graduate School Program (Hanyang University) and No.2021-0-01343,
Artificial Intelligence Graduate School Program (Seoul National University)).

\bibliographystyle{acl_natbib}
\bibliography{emnlp2021}

\appendix

\section{Appendix}

\subsection{List of Non-English Monolingual PLMs} \label{subsec: List of Non-English Monolingual PLMs}

The PLMs we utilize per language are listed as follows.
\textbf{German:} bert-base-german (\url{https://deepset.ai/german-bert}).
\textbf{French:} camembert \cite{martin2019camembert}.
\textbf{Swedish:} bert-base-swedish (\url{https://github.com/huggingface/transformers/tree/master/model_cards/KB/bert-base-swedish-cased}).
\textbf{Korean:} KoBERT-base (\url{https://github.com/SKTBrain/KoBERT}).

\end{document}